\documentclass[runningheads]{llncs}

\usepackage{eccv}
\usepackage{eccvabbrv}
\usepackage{graphicx}
\usepackage{booktabs}
\usepackage{amsmath, amssymb, bm}

\usepackage[accsupp]{axessibility}

\usepackage[pagebackref,breaklinks,colorlinks,citecolor=eccvblue]{hyperref}

\usepackage{orcidlink}

\begin{document}
	
	\title{Generic Camera Calibration using Blurry Images} 
	
	\author{Zezhun Shi}
	
	\institute{Independent Researcher
		\email{wsfddn@foxmail.com}}
	
	\maketitle

	\begin{abstract}
		Camera calibration is the foundation of 3D vision. Generic camera calibration can yield more accurate results than parametric camera calibration. However, calibrating a generic camera model using printed calibration boards requires far more images than parametric calibration, making motion blur practically unavoidable for individual users. As a first attempt to address this problem, we draw on geometric constraints and a local parametric illumination model to simultaneously estimate feature locations and spatially varying point spread functions, while resolving the translational ambiguity that need not be considered in conventional image deblurring tasks. Experimental results validate the effectiveness of our approach.
		
	\end{abstract}
	
	\section{Introduction}
	\label{sec:intro}
	
	\begin{figure}[tb]
		\centering
		\includegraphics[width=0.6\linewidth]{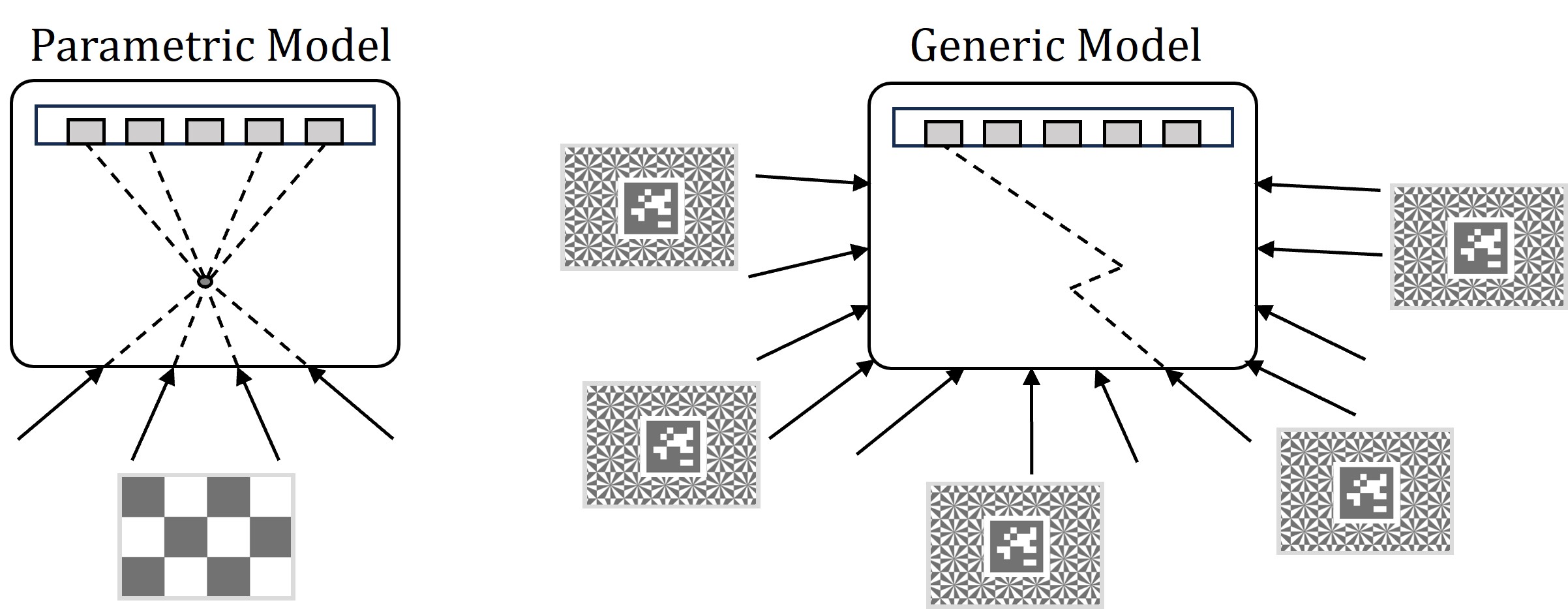}
		\caption{Parametric model and generic model. The latter can yield more accurate results than former but requires far more calibration images.}
		\label{fig:related}
	\end{figure}
	
	Camera calibration establishes the mapping from 3D observation rays to image pixels and is fundamental to 3D vision~\cite{Orb3,PatchmatchStereo,SGM,Nerf,3DGS}.
	Camera models fall into two categories: \emph{parametric} models~\cite{Brown1971,Brown1986,Zhang2000,fisheye1,fisheye2} that assume a global functional form for lens distortion, and \emph{generic} models~\cite{Grossberg2001,Sturm2004,GVC1,Bspline,Schops2020} that independently calibrate each ray without such assumptions.
	Sch\"ops~\etal~\cite{Schops2020} showed that generic models eliminate systematic directional bias inherent in parametric models, benefiting downstream tasks such as stereo depth estimation~\cite{SGM,PatchmatchStereo}.
	
	All existing calibration methods assume sharp input images.
	This is easily satisfied for parametric calibration, which needs only a dozen carefully captured images.
	Generic calibration, however, may require thousands of images~\cite{Schops2020} to cover the full pixel grid, making it practically impossible to guarantee the absence of motion blur throughout capture.
	Inexpensive cameras with low frame rates further aggravate this problem.
	Discarding blurry frames wastes valuable pixel coverage and prolongs data collection.
	
	\begin{figure}[h!]
		\centering
		\includegraphics[width=1\linewidth]{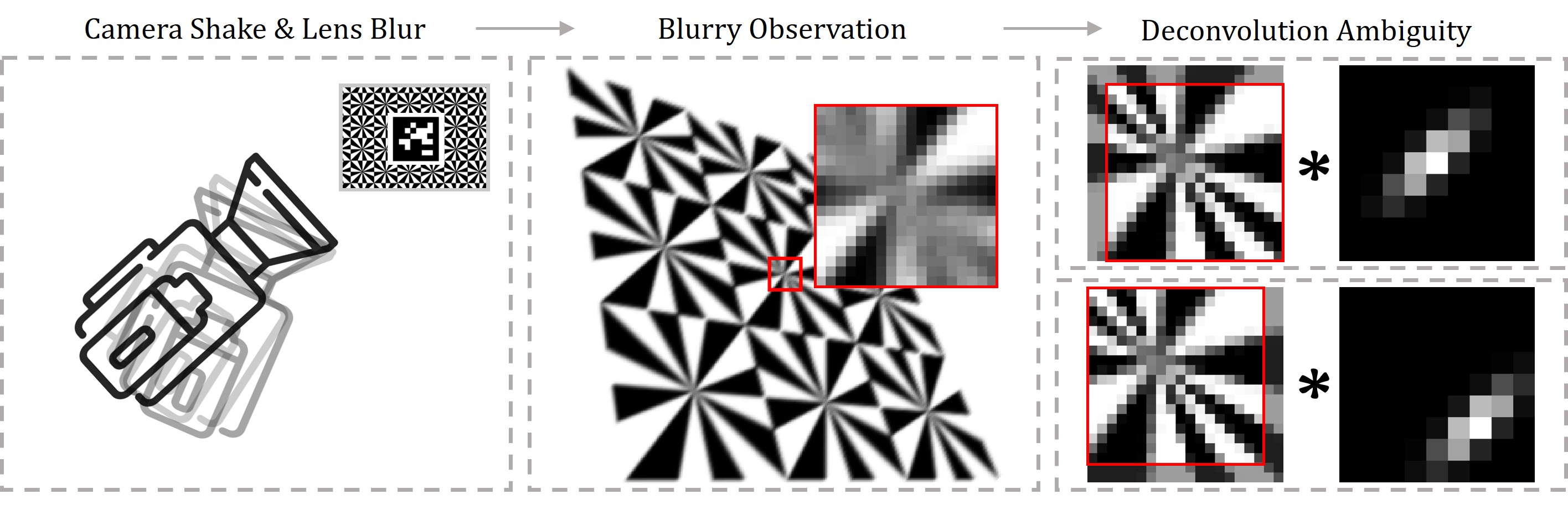}
		\caption{Generic camera calibration requires far more images than parametric calibration, making motion blur practically unavoidable. Standard deconvolution recovers visually plausible images but leaves feature positions ambiguous up to an arbitrary translation due to the shift equivariance of convolution.}
		\label{fig:intro}
	\end{figure}
	
	Applying general deblurring~\cite{Fergus2006,Michaeli2014,DeepDeblur1,DeepDeblur2,DeepDeblur3} as preprocessing cannot achieve the subpixel geometric fidelity that calibration demands.
	Existing PSF estimation methods using calibration patterns~\cite{Delbracio2012,PSFDynamic,PSFCheckerboard} assume a spatially invariant PSF per region and require that the pattern to image mapping is already known from a separate feature extraction step.
	Under motion blur, this prerequisite fails: conventional feature detectors cannot localize features in blurred images, creating a circular dependency.
	Furthermore, a fundamental issue arises when incorporating deconvolution into the geometric pipeline.
	Due to the shift equivariance of convolution, any translational error in the recovered latent image is absorbed by the kernel as an equal and opposite displacement (Fig.~\ref{fig:intro}).
	While this \emph{translational ambiguity} is harmless for standard deblurring where only visual quality matters, it directly corrupts the geometric accuracy of extracted features for calibration.
	
	We propose a framework that simultaneously estimates feature positions and spatially varying PSFs from blurry images, enabling generic camera calibration without requiring blur free capture.
	Our key idea is to parameterize the latent image in each local region as a \emph{homography} acting on the known calibration pattern with a linear illumination correction, reducing the latent image from tens of thousands of free pixel values to only 14 parameters.
	This makes the joint estimation of image content and blur kernel well conditioned while directly yielding the geometric mapping needed for calibration.
	Since neighboring blocks share pattern vertices, the homographies of adjacent blocks are geometrically coupled at their boundaries. We exploit this coupling to enforce inter block consistency, enabling spatially varying PSF estimation without the computational burden of global deconvolution~\cite{VariantPSF1,VariantPSF2,VariantPSF3}.
	The remaining global translational ambiguity is eliminated by aligning deconvolved features to a parametric camera calibrated from a small set of sharp images.
	
	Our contributions are as follows:
	\begin{itemize}
		\item We formulate a \emph{homography parameterized local deconvolution} that jointly estimates the geometric mapping and blur kernel from a known calibration pattern, breaking the circular dependency between feature extraction and deblurring. We further derive a fully differentiable approximation of the star shaped calibration pattern for PSF estimation, enabling gradient based optimization of the homography parameters (see supplementary material).
		\item We introduce \emph{geometric inter block constraints} that couple adjacent homographies through shared pattern vertices, enabling spatially varying PSF estimation. Unlike prior methods~\cite{Delbracio2012} that can only handle a single invariant PSF per region, our formulation naturally accommodates spatial variation of both optical and motion blur.
		\item We address the translational ambiguity inherent in the deconvolution process, which would otherwise corrupt geometric accuracy, and resolve it through local geometric alignment across neighboring blocks combined with global parametric camera alignment.
	\end{itemize}

	\section{Related Work}
	\noindent \textbf{Camera calibration} fits a math model to a calibration target with known 3D structure and its image mapping. Traditionally, camera models apply radial and tangential corrections to the pinhole camera with few parameters, mainly including Brown-Conrady\cite{Brown1971,Brown1986,Zhang2000,CalibrationOther} and fisheye cameras\cite{fisheye1,fisheye2}, called parametric or global model\cite{Surface_Model}. One core operation for parametric camera model calibration is localizing sparse calibration target features such as checkerboards\cite{CornerSubpix,Rochade,Duda2018} or circles\cite{Heikkila2000,circle2021,Song2024} in images with subpixel algorithms. In practice, defocus often causes imprecise feature point localization. To address this, many researchers use phase patterns that are more robust to defocus instead of traditional static calibration targets\cite{phase_simple,phase_sin,phase_net}. Other attempts include Ha~\etal \cite{Ha15}using complementary binary patterns, modeling defocus as a Gaussian function and obtaining feature points through deconvolution.
	
	Parametric models satisfy most calibration requirements. However, considering the existence of non-perspective imaging systems such as curved mirrors and meniscus lenses, Grossberg and Nayar proposed a different camera model that independently calibrates the mapping between each ray and pixel\cite{Grossberg2001}(Fig.~\ref{fig:related}), with subsequent research called ray model, surface model, or generic model\cite{Surface_Model}. In their experiments, they displayed binary fringes on two screens at known distances to obtain physical rays corresponding to camera pixels. Although they achieved meaningful results, the resolution of binary fringes and high precision screen placement brought limitations in calibration accuracy and operational convenience. Sturm and Ramalingam algebraically provided a method to estimate camera pose under generic models with arbitrarily placed calibration targets\cite{Sturm2004}, removing the constraint of fixed screen placement. Dunne~\etal used phase patterns that provide dense features\cite{GVC1}. They first calibrated ray mappings through a display to obtain a virtual central camera, then used conventional procedures to calibrate the pinhole camera, simplifying generic camera calibration complexity. This method was extended by Brousseau and Roy to combinations of multiple virtual central cameras\cite{GVC2}, achieving calibration of large field of view axial fisheye cameras. Although phase patterns provide dense ray mappings, the calibration procedure is less convenient than traditional printed calibration targets. Beck and Stiller used B-spline interpolation to fit generic models on discrete grids\cite{Bspline}, achieving high precision generic model calibration using only checkerboards. Building on this, Schops~\etal used star-shaped calibration targets that provide more image gradient information than checkerboards, along with new subpixel algorithms and a complete BA pipeline\cite{Schops2020}. In their experiments, this complete generic camera calibration pipeline demonstrated extremely high precision. More importantly, they showed that parametric camera models have systematic reprojection directional errors, so even small subpixel calibration precision improvements in generic models can benefit vision applications relying on continuous image information, such as stereo depth estimation\cite{SGM,PatchmatchStereo}.
	
	\noindent \textbf{PSF estimation} is fundamental to both optics and computer vision. The most direct approach samples the response of a point light source, but this is highly sensitive to noise. More robust methods employ artificial calibration patterns with known structure, such as one dimensional fringes~\cite{PSF1D}, checkerboards~\cite{PSFCheckerboard}, segmented arcs~\cite{Joshi2008,PSFArc}, and circles~\cite{Kee2011}. Delbracio~\etal~\cite{Delbracio2012} proved that Bernoulli noise patterns can approach optimal estimation precision. A common prerequisite of all these methods is that the mapping from the ideal pattern to its image must be known beforehand, typically obtained through a separate feature extraction procedure. Mosleh~\etal~\cite{PSFDynamic} used a display to show two patterns in sequence, one for geometric localization and another for PSF estimation. However, these approaches estimate the PSF only at isolated locations and fixed depths. To obtain spatially dense estimates, one can parameterize the PSF, \eg, with 2D Gaussians~\cite{Kee2011} or the Sinh Arcsinh model that better captures asymmetric aberrations~\cite{PSFSinAsin}. Extending estimation across the full depth range requires calibration images at multiple distances; Shih~\etal~\cite{PSFSimu} combined point source sampling with a calibrated lens model to reduce the required captures by roughly two orders of magnitude. A key limitation shared by all the above methods is the assumption that the pattern to image mapping is already precisely known. In our problem, motion blur prevents conventional feature extraction from providing this mapping, motivating our joint estimation formulation.
	
	For image deblurring, the goal shifts from precise PSF recovery to producing latent images that are visually pleasing. Blind motion deblurring, where no calibration pattern is available, is typically addressed through alternating optimization over the latent image and the kernel. The choice of latent image prior is critical. Fergus~\etal~\cite{Fergus2006} learned mixture of Gaussians priors on natural image gradients, enabling blind deconvolution of single images with large blur for the first time. Michaeli and Irani~\cite{Michaeli2014} showed that internal patch recurrence within a single image can serve as a more robust prior. These ideas inspired subsequent learning based deblurring methods~\cite{DeepDeblur1,DeepDeblur2,DeepDeblur3}. For domain specific tasks, tailored priors can be far more effective: S\"or\"os~\etal~\cite{QRPSF} exploited the binary structure of QR codes, achieving faster and more accurate deblurring than general purpose algorithms. Our work follows this philosophy of exploiting domain structure, but goes further by parameterizing the latent image as a homography on a known calibration pattern, which simultaneously yields the geometric mapping needed for calibration.
	
	All the above methods assume a spatially invariant blur kernel, which does not hold under general camera motion or across the field of view. For spatially varying PSF problems, Nagy and O'Leary~\cite{VariantPSF1} proposed image domain interpolation, Flicker and Rigaut~\cite{VariantPSF2} used multiple PSF patterns, and Hirsch~\etal~\cite{VariantPSF3} showed that applying spatially varying weights before and after convolution yields superior results. However, these methods must jointly solve for all pixel values and all local kernels, incurring prohibitive computational cost and yielding severely underdetermined systems that limit the number of PSF regions. In contrast, our homography parameterization reduces each local region to 14 unknowns and introduces geometric coupling between neighboring blocks through shared calibration pattern vertices, enabling spatially varying PSF estimation without global deconvolution.
	
	\section{Preliminary}
	
	\subsection{Brown Conrady model}
	A pinhole camera with lens distortion maps a 3D point $P$ to image point $p$ via
	\begin{equation}
		p = \pi_{d}(\mathbf{M}, P) = \mathbf{K} \begin{pmatrix} \mathbf{d}\!\left([\mathbf{M} P]_{\sim}\right) \\ 1 \end{pmatrix},
		\label{Brown_Conrady}
	\end{equation}
	where $\mathbf{M} = [\mathbf{R} \mid \mathbf{t}]$ is the camera extrinsic matrix, $\mathbf{K}$ is the intrinsic matrix, $[\cdot]_{\sim}$ denotes the conversion from homogeneous to normalized image coordinates, i.e., $(X, Y, Z)^\top \mapsto (X/Z,\, Y/Z)^\top$, and $(\cdot\,;\, 1)^\top$ converts back to homogeneous form. The function $\mathbf{d}: \mathbb{R}^2 \to \mathbb{R}^2$ applies lens distortion following the Brown\textendash Conrady model:
	\begin{equation}
		\mathbf{d}(\mathbf{x}) = (1 + k_1 r^2 + k_2 r^4 + k_3 r^6)\mathbf{x} + \begin{pmatrix} 2p_1 x y + p_2(r^2 + 2x^2) \\ p_1(r^2 + 2y^2) + 2p_2 x y \end{pmatrix},
	\end{equation}
	where $\mathbf{x} = (x, y)^\top$ is the normalized image coordinate and $r^2 = x^2 + y^2$.
	
	\subsection{Homography and Camera Motion}
	For a planar calibration target, the homography $\mathbf{H} \in \mathbb{R}^{3 \times 3}$ relating the pattern plane to the (undistorted) image is determined by the camera motion as
	\begin{equation}
		\mathbf{H} = \mathbf{K}(\mathbf{r}_1 \mid \mathbf{r}_2 \mid \mathbf{t}),
		\label{eq:T}
	\end{equation}
	where $\mathbf{r}_1, \mathbf{r}_2$ are the first two columns of $\mathbf{R}$. A 2D translation is a special homography
	\begin{equation}
		\mathbf{T}(\mathbf{x}) = \begin{pmatrix} 1 & 0 & x_1 \\ 0 & 1 & x_2 \\ 0 & 0 & 1 \end{pmatrix}.
	\end{equation}
	
	\subsection{Translational Ambiguity in Deconvolution}
	Under spatially invariant blur, the observed image is modeled as
	\begin{equation}
		I = \mathbf{k} * L + n,
	\end{equation}
	where $L$ is the latent sharp image, $\mathbf{k}$ is the blur kernel, $*$ denotes convolution, and $n$ represents additive noise~\cite{Delbracio2012}. Deconvolution aims to recover $L$ and/or $\mathbf{k}$ from $I$. However, due to the shift equivariance of convolution, for any 2D displacement $\boldsymbol{\delta}$, let $L'(\mathbf{x}) = L(\mathbf{x} - \boldsymbol{\delta})$ and $\mathbf{k}'(\mathbf{x}) = \mathbf{k}(\mathbf{x} + \boldsymbol{\delta})$, then
	\begin{equation}
		\mathbf{k}' * L' = \mathbf{k} * L,
	\end{equation}
	meaning that any translational error in the recovered latent image is absorbed by the kernel as an equal and opposite displacement. This ambiguity makes the decomposition non-unique without additional constraints. 
	
	\section{Method}
	\begin{figure}[h]
		\centering
		\includegraphics[width=1\linewidth]{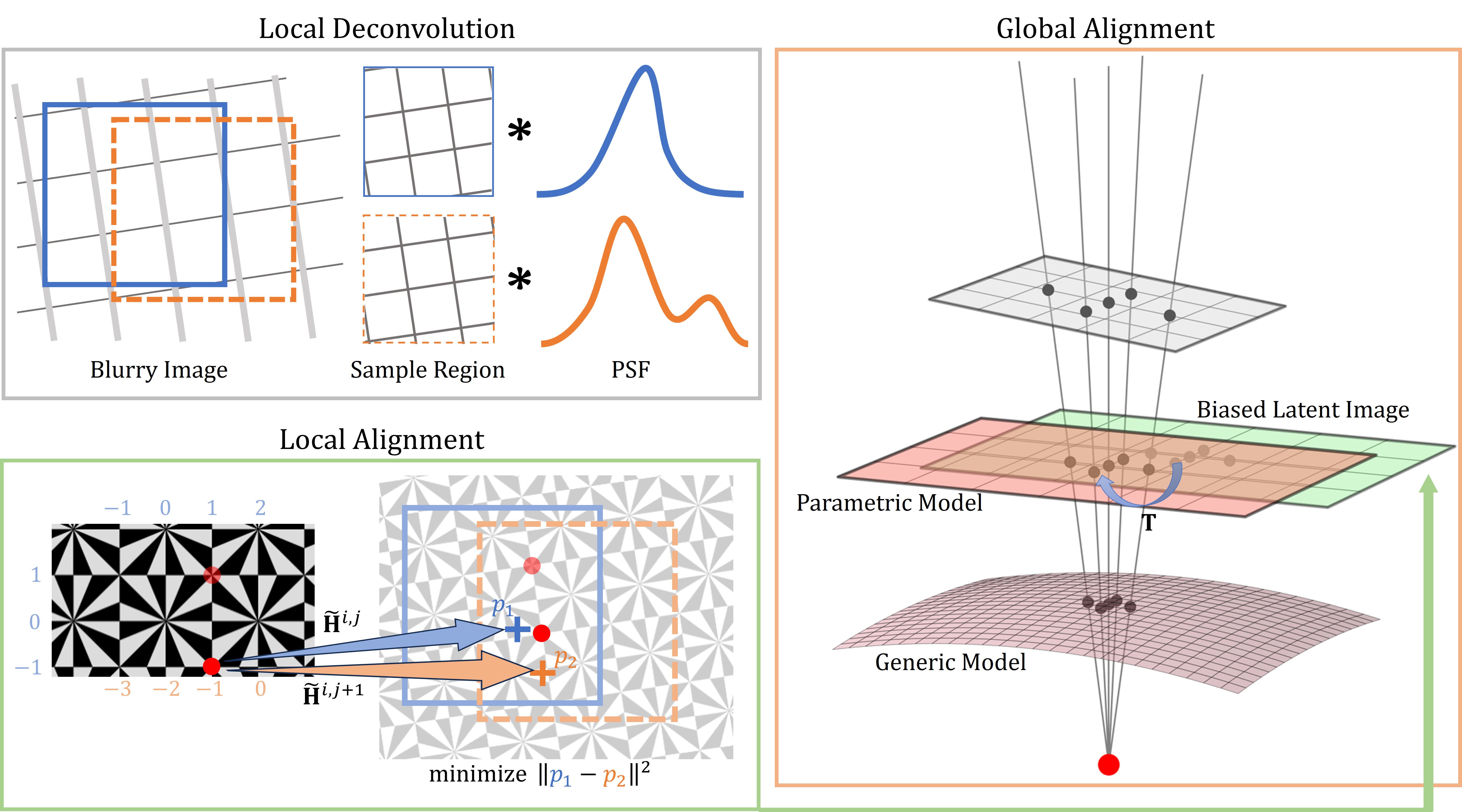}
		\caption{
			Overview of the proposed method. We use local homographies and illumination parameters to replace classical feature based global mapping to describe the latent image, with geometric constraints enforcing inter-block consistency. Global translational ambiguity is resolved by alignment with a calibrated parametric camera.
		}
		\label{fig:method}
	\end{figure}
	
	\subsection{Overview}
	Our objective is to extract subpixel precision feature points for camera calibration, a task that remains challenging for conventional deconvolution algorithms. Two primary obstacles exist: first, the inherent translational ambiguity in the deconvolution process as illustrated in Figure \ref{fig:intro}; second, the complexity of the blur kernel which conflates lens and motion blur in a spatially varying manner. Rather than performing heavy global deconvolution, we employ a local deconvolution strategy integrated with local region homography estimation. We utilize the calibration pattern proposed by Schops \etal \cite{Schops2020}, featuring a central AprilTag surrounded by star pattern elements. Initial homographies are derived from the central AprilTags and expanded via a region growing procedure. To resolve translational ambiguity, we align each frame to a calibrated parametric camera through a multi stage alignment process.
	
	\subsection{Local Deconvolution}
	PSF estimation and image deblurring typically involve minimizing a convolution cost function. In our framework, we avoid explicit feature extraction algorithms which often fail under motion blur. Instead, we represent the local image as a homography acting on the ideal star pattern with a linear varying brightness correction. The optimization objective is formulated as:
	\begin{equation}
		\arg\min_{\mathbf{H}, \mathbf{k}, \mathbf{p}} \|I - \mathbf{k} * (S(\mathbf{H}) \odot A(\mathbf{p}) + B(\mathbf{p}))\|^2+\lambda\|\mathbf{k}\|^2,
		\label{eq:deconvolution}
	\end{equation}
	where $I$ is the observed image, $\mathbf{k}$ is the blur kernel, $S(\mathbf{H})$ is the ideal star pattern warped by homography $\mathbf{H}$, and $\lambda$ is a regularization weight. Spatially varying amplitude $A(\mathbf{p})$ and bias $B(\mathbf{p})$ are defined using normalized pixel coordinates $(u, v)$ as:
	\begin{equation}
		A(\mathbf{p}) = a_1 u + a_2 v + a_3, \quad B(\mathbf{p}) = b_1 u + b_2 v + b_3,
	\end{equation}
	where $\mathbf{p} = [a_1, a_2, a_3, b_1, b_2, b_3]^\top$. 
	
	We implement this optimization within a differentiable PyTorch framework. For each iteration, the kernel $\mathbf{k}$ and brightness parameters $\mathbf{p}$ are solved analytically using the closed form least squares solution. Given fixed $\mathbf{k}$ and $\mathbf{H}$, we define basis images $\mathbf{f}_i = \mathbf{k} * \phi_i$, where $\boldsymbol{\phi} = [S u, S v, S, u, v, \mathbf{1}]$. The optimal brightness parameters are computed as $\mathbf{p}^* = (F^\top F)^{-1} F^\top \mathrm{vec}(I)$, where $F = [\mathrm{vec}(\mathbf{f}_1), \ldots, \mathrm{vec}(\mathbf{f}_6)] \in \mathbb{R}^{N \times 6}$ is the design matrix composed of vectorized basis images. The 8 dimensional homography matrix $\mathbf{H}$ is optimized using gradient descent with backpropagation, allowing for precise convergence in a continuous parameter space.
	
	\subsection{Local Alignment}
	Deconvolution results between adjacent blocks may exhibit overlapping discrepancies due to the shift invariant property of convolution. We introduce a translation correction $\mathbf{T}$ to the estimated homography matrices:
	\begin{equation}
		\tilde{\mathbf{H}}^{i,j}(\mathbf{x_L}^{i,j}) = \mathbf{T}(\mathbf{x_L}^{i,j}) \mathbf{H}^{i,j},
	\end{equation}
	where $i, j$ denote the grid indices and $\mathbf{x_L}$ represents the 2D translation vector. For each element block, we consider its 4 neighbors and minimize the distance between shared vertices:
	\begin{equation}
		\begin{aligned}
			\arg \min _{\mathbf{x_L}} \sum_{i, j, m} \Bigg\{ &d\left[\tilde{\mathbf{H}}^{i, j}(\mathbf{x_L}^{i, j})\begin{pmatrix}
				1 \\ m \\ 1
			\end{pmatrix}, \tilde{\mathbf{H}}^{i, j+1}(\mathbf{x_L}^{i, j+1})\begin{pmatrix}
				-1 \\ m \\ 1
			\end{pmatrix}\right]^2 \\
			+ &d\left[\tilde{\mathbf{H}}^{i, j}(\mathbf{x_L}^{i, j})\begin{pmatrix}
				m \\ 1 \\ 1
			\end{pmatrix}, \tilde{\mathbf{H}}^{i+1, j}(\mathbf{x_L}^{i+1, j})\begin{pmatrix}
				m \\ -1 \\ 1
			\end{pmatrix}\right]^2 \Bigg\} 
		\end{aligned},
	\end{equation}
	where $m \in \{-1, 1\}$ and $d[\cdot,\cdot]$ is the Cartesian distance. To prevent the accumulation of drift, we monitor the intensity weighted centroid of the PSF and apply a corrective translation to recenter the PSF after the initial estimation stage, ensuring the optimization remains stable.
	
	\subsection{Global Alignment}
	\label{sec:global_alignment}
	Parametric camera models often fail to capture systematic reprojection directional bias. To resolve the translational ambiguity inherent in deconvolution, we align the observations to a calibrated parametric model. The discrepancy between an observation $p^{i,j}$ and its parametric projection $\pi_d(\mathbf{M}, P^{i,j})$ is modeled as:
	\begin{equation}
		p^{i,j} - \pi_d(\mathbf{M}, P^{i,j}) = \epsilon_{\text{deconv}}^{i,j} + \epsilon_{\text{align}}^{i,j} + \mathbf{b}(\mathbf{r}^{i,j}),
		\label{eq:error_decomposition}
	\end{equation}
	where $\pi_d$ denotes the Brown Conrady model, $P^{i,j}$ is the 3D point, and $\mathbf{r}^{i,j}$ is the ray direction. The terms $\epsilon_{\text{deconv}}$ and $\epsilon_{\text{align}}$ are stochastic errors from deconvolution and local alignment respectively, and $\mathbf{b}(\mathbf{r})$ represents the deterministic systematic bias. Under the assumption that stochastic terms are unbiased with near zero mean, their influence vanishes upon joint optimization over multiple observations, leaving the systematic bias as the primary residual. 
	
	We formulate the global alignment as a robust optimization problem to solve for camera motion $\mathbf{M}$ and global translation $\mathbf{x_G}$:
	\begin{equation}
		\arg\min_{\mathbf{x_G},\mathbf{M}}\sum_{i, j}\rho\left(d\left[\pi_{d}(\mathbf{M},P^{i,j}),\mathbf{T}(\mathbf{x_G})p^{i,j}\right]^2\right),
		\label{eq:global_align}
	\end{equation}
	where $\rho$ denotes a robust kernel. 
	
	\paragraph{Spatially varying bias compensation.}
	In practice, we observe that the residual errors after global alignment are not uniformly distributed but exhibit a distinct spatial structure across the image plane. Initial experiments showed that a single global translation $\mathbf{x_G}$ is insufficient to account for localized PSF estimation biases. While partitioning the grid into four independent quadrants and solving for separate translation offsets can partially mitigate these variations, this discrete approach introduces artificial jump discontinuities at the quadrant boundaries. To address these residual structures without sacrificing the smoothness of the correction, we model the bias as a continuous bilinear field $\hat{\mathbf{b}}(\bar{i}, \bar{j})$ over normalized grid coordinates $(\bar{i}, \bar{j}) \in [-1, 1]^2$:
	\begin{equation}
		\hat{\mathbf{b}}(\bar{i}, \bar{j}) = \mathbf{a}_0 + \mathbf{a}_1 \bar{i} + \mathbf{a}_2 \bar{j} + \mathbf{a}_3 \bar{i}\bar{j},
		\label{eq:bilinear_bias}
	\end{equation}
	where $\mathbf{a}_k \in \mathbb{R}^2$ are coefficient vectors. The alignment then proceeds by alternating between two steps. First, given the current bias field, the observed points are corrected and the camera pose $\mathbf{M}$ is updated via PnP. Second, given the pose, we compute the per point residuals and update the bias coefficients $\{\mathbf{a}_k\}$ through weighted least squares. This iterative process ensures a rigid global constraint while smoothly compensating for localized spatial trends.
	
	\section{Experiments}
	
	\subsection{Patterns Comparison}
	\begin{figure}[h!]
		\centering
		\includegraphics[width=1\linewidth]{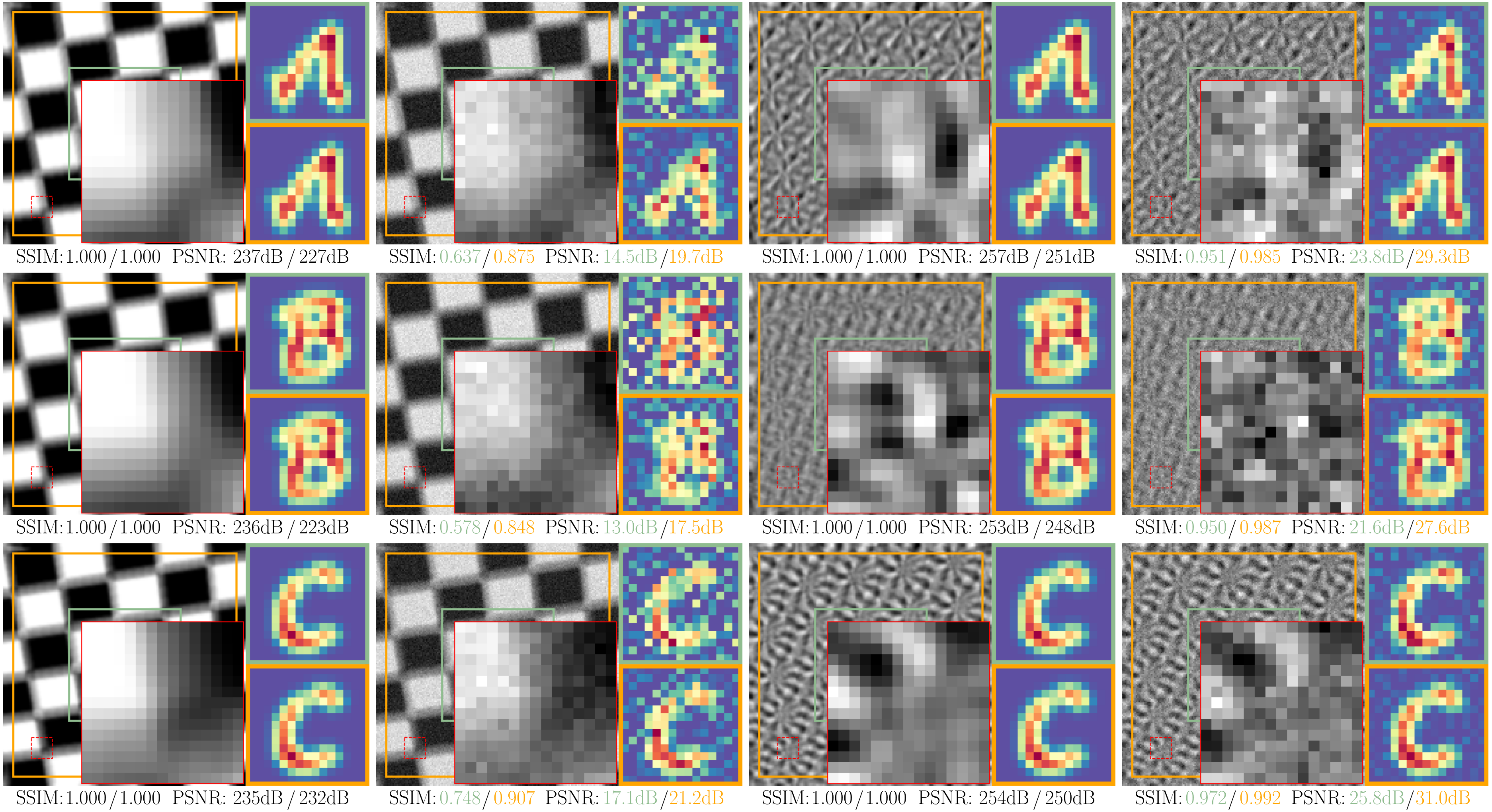}
		\caption{Deconvolution results for four PSF glyphs under checkerboard and star calibration patterns, with and without 5\% Gaussian noise. Each group shows an $80\times80$ and $160\times160$ px crop with SSIM/PSNR. The star pattern is significantly more robust to noise.}
		\label{fig:exp2_test}
	\end{figure}
	We quantitatively assess the accuracy of PSF estimation using two distinct calibration patterns: a conventional checkerboard (featuring two intersecting edge directions) and the star shaped pattern (offering eight intersecting edge directions). To evaluate the generalization of our PSF recovery, we employ a Lucida Handwriting glyph as a complex blur kernel, which is color inverted, resized to $15\times15$ pixels, and smoothed with a Gaussian filter ($\sigma=0.8$). This glyph serves as the sole unknown in Eq.~\ref{eq:deconvolution} while all other geometric parameters remain fixed.
	
	As demonstrated in Figure~\ref{fig:exp2_test}, both patterns achieve comparable reconstruction quality under noise free conditions. However, when subjected to 5\% Gaussian noise, the checkerboard pattern exhibits severe degradation, with SSIM falling to approximately $0.58$ and PSNR dropping to $13$ dB. This failure stems from the limited edge orientations of the checkerboard, which leaves the PSF estimation underdetermined along diagonal directions and highly sensitive to stochastic noise perturbations. In contrast, the star shaped pattern provides dense frequency domain coverage through its eight directional edges, maintaining robust performance with an SSIM of approximately $0.96$ and PSNR above $22$ dB. This superior noise robustness and geometric stability confirm that the star pattern is the optimal choice for high precision PSF calibration in blurry environments.

	\subsection{Global Alignment Verification}
	\label{sec:global_align_verify}
	
	\begin{figure}[h]
		\centering
		\includegraphics[width=0.995\linewidth]{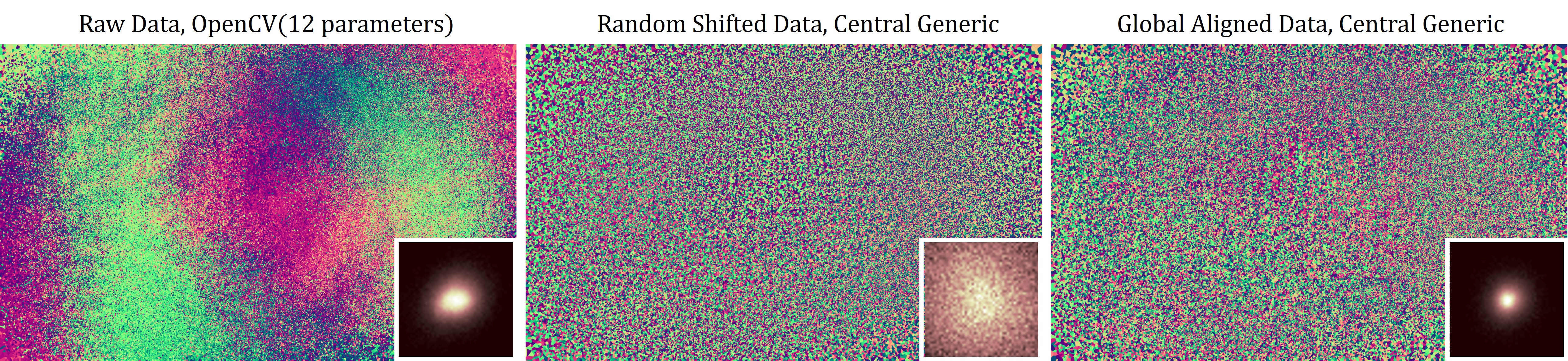}
		\caption{Distribution of reprojection direction (hue) and magnitude.}
		\label{fig:exp42}
	\end{figure}
	
	\begin{figure}[h]
		\centering
		\includegraphics[width=1\linewidth]{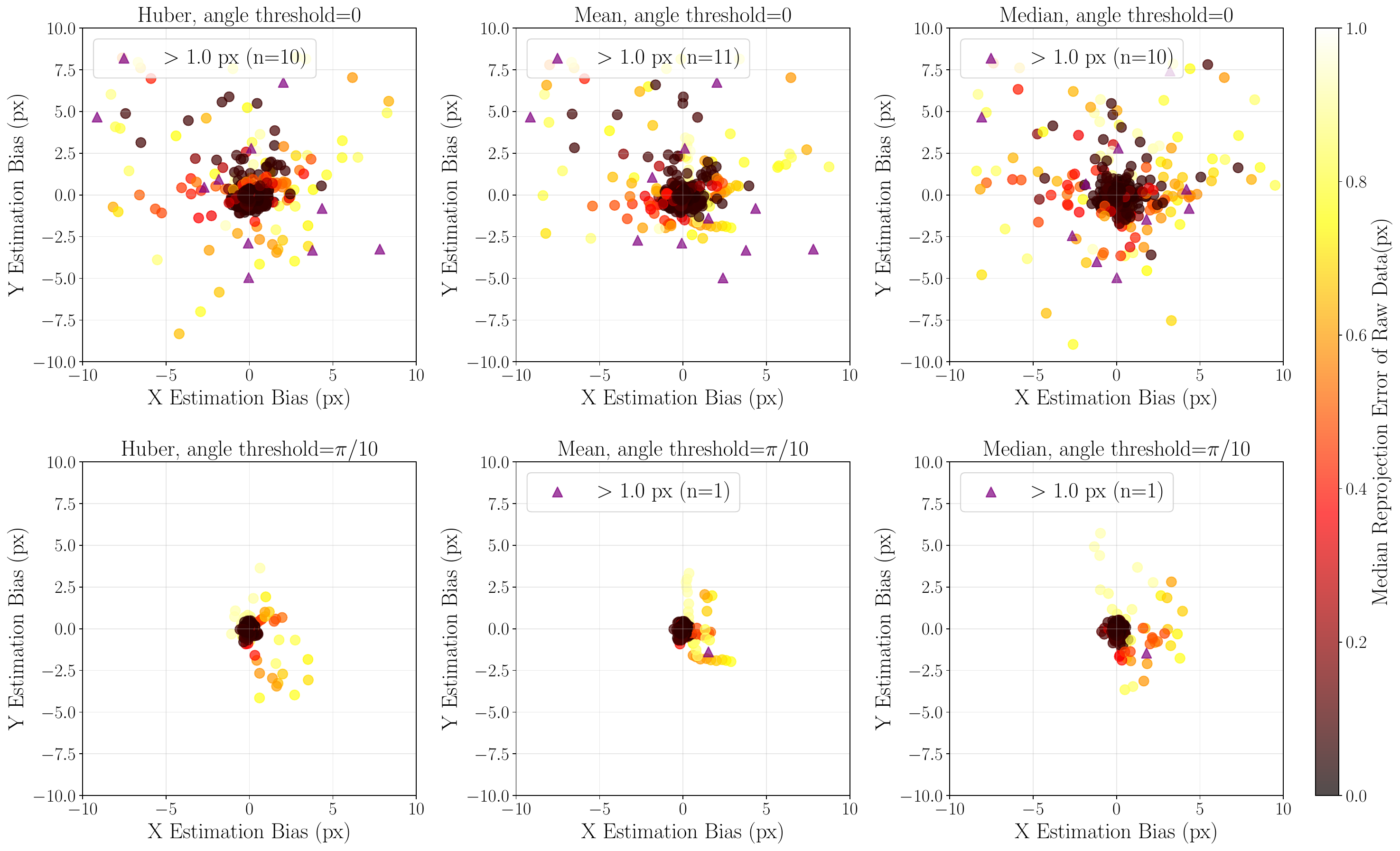}
		\caption{Transition bias distribution against median reprojection error of raw data (per frame).}
		\label{fig:exp41}
	\end{figure}
	
	To validate the effectiveness of our global alignment formulation, we conduct controlled experiments using the D435I calibration dataset provided by Schops \etal \cite{Schops2020}. We extract ground truth features and introduce synthetic random translations within a range of $\pm 5$ pixels to each frame, simulating the translational ambiguity inherent in the deconvolution process. Using 20 uniformly sampled images, we calibrate a baseline parametric model $\pi_d$ and 3D points $P$ to serve as the alignment reference.
	
	We evaluate the performance of three loss functions: the Huber robust kernel, the arithmetic mean, and the median. As illustrated in Fig.\ref{fig:exp42}, our alignment method significantly reduces the introduced random offsets, yielding a uniform distribution of reprojection directions. We observe that when the camera z axis is nearly parallel to the target normal, the perspective transformation approaches an affine mapping, rendering 2D translation estimation ill conditioned. To mitigate this, we apply an orientation filter based on the angle between the optical axis and the target normal. As shown in Fig.\ref{fig:exp41}, filtering frames below a threshold of $\pi/10$ eliminates large estimation biases (represented by black dots) and establishes a strong correlation between the residual alignment error and the original feature noise. Numerical results in Tab.\ref{tab:exp4} confirm that the Huber loss achieves the highest precision, reaching an alignment error of $0.042$ px. This demonstrates that our alignment strategy effectively recovers the deterministic systematic bias while minimizing the influence of stochastic feature noise.
	
	\begin{table}[h]
		\centering
		\caption{Medium/mean reprojection errors comparison (px).}
		\resizebox{1\linewidth}{!}{%
			\setlength{\tabcolsep}{8pt}
			\begin{tabular}{c|ccc|ccc}
				\hline
				Angle Threshold& \multicolumn{3}{c|}{0} & \multicolumn{3}{c}{$\pi$/10} \\
				\hline
				Loss Type& Huber & Mean & Median & Huber & Mean & Median \\
				\hline
				Random Shifted & 0.257/0.275 & 0.259/0.278 & 0.260/0.278 & 0.236/0.279 & 0.243/0.286 & 0.243/0.286 \\
				Global Aligned & \textbf{0.061}/\textbf{0.064} & 0.064/0.068 & 0.090/0.096 & \textbf{0.042}/\textbf{0.044} & 0.058/0.059 & 0.065/0.066 \\
				Random Shifted(self) & 0.213/0.278 & 0.213/0.278 & 0.213/0.278 & 0.289/0.354 & 0.288/0.354 & 0.288/0.354 \\
				Global Aligned(self) & \textbf{0.036}/\textbf{0.050} & 0.037/0.051 & 0.039/0.057 & \textbf{0.036}/\textbf{0.047} & 0.037/0.048 & 0.039/0.052 \\
				Raw & 0.028/0.036 & 0.028/0.036 & 0.028/0.036 & 0.031/0.040 & 0.031/0.039 & 0.031/0.039 \\
				\hline
			\end{tabular}%
		}
		\label{tab:exp4}
	\end{table}
	
	\subsection{Real World Evaluation}
	We evaluate the complete pipeline on real data captured with an Intel RealSense D435I (global shutter, $1280\times720$) using the star pattern calibration target of Sch\"{o}ps~\etal~\cite{Schops2020}. We capture a blurry set of 204 frames at 15\,fps with deliberate hand shake, and a small sharp set of 20 images for calibrating the parametric camera model $\pi_d$ and 3D target points $P$ used in global alignment (Eq.~\ref{eq:global_align}). We verify that the D435I sensor response is linear ($R^2 > 0.999$) across the operating intensity range, confirming that no radiometric calibration is needed for deconvolution despite the availability of tools such as \texttt{cv::CalibrateDebevec} in OpenCV~\cite{CalibrateDebevec}.
	
\begin{figure}[h!]
		\centering
		\includegraphics[width=1\linewidth]{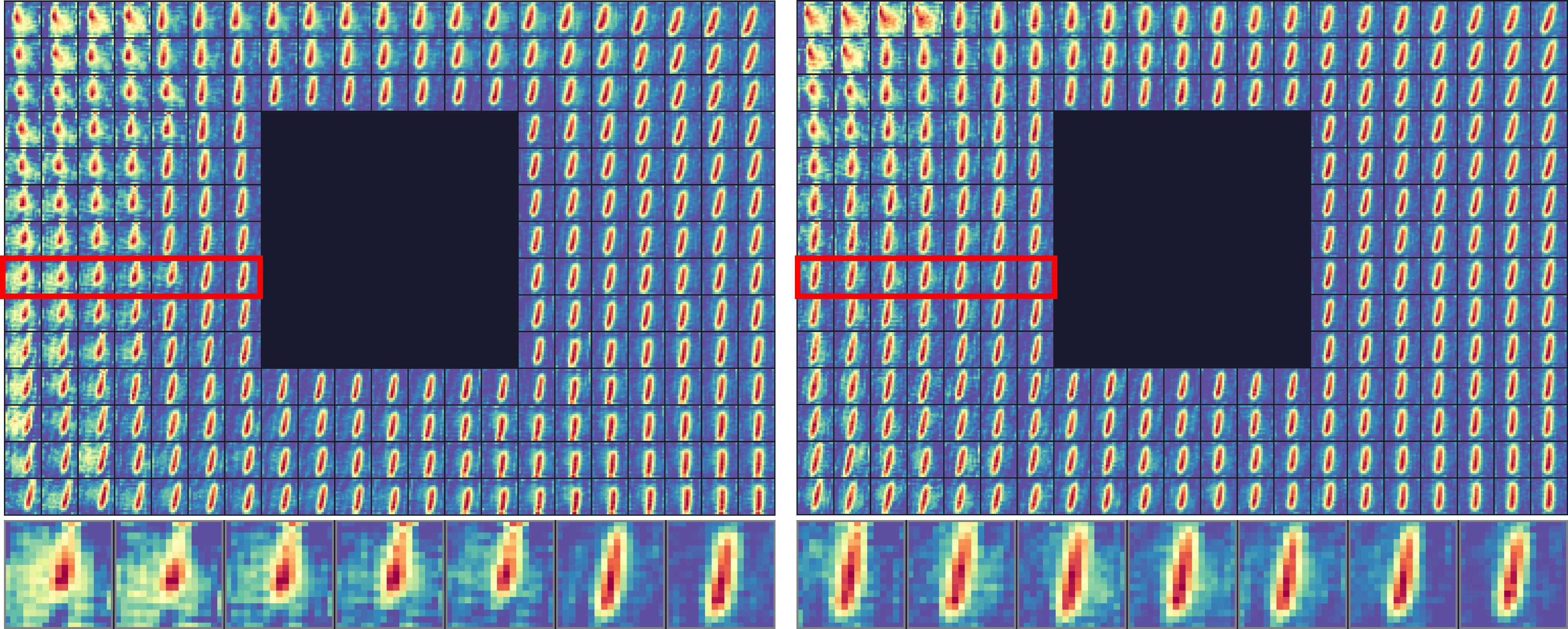}
		\caption{Estimated spatially varying PSF field for a sample blurry image. Left: without centroid correction. Right: with corrective recentering, showing spatially coherent motion blur that varies smoothly across the field. Bottom row: magnified PSFs from the red boxed region.}
		\label{fig:exp_real1}
	\end{figure}

	\paragraph{PSF estimation and centroid correction.}
	We apply our local deconvolution (Eq.~\ref{eq:deconvolution}) to the 204 blurry frames with a PSF window of $17\times17$ pixels. Fig.~\ref{fig:exp_real1} shows the estimated spatially varying PSF field for a representative frame. Without centroid correction (left), accumulated translational drift causes PSFs near the image boundary to shift off center and become irregular. With corrective recentering (right), the PSF estimates are spatially coherent with a consistent motion blur direction that varies smoothly across the field. We pre filter frames whose center region median $\sigma_{\mathrm{major}} < 3.8$\,px, as these contain insufficient blur for reliable PSF estimation, retaining 42 blurry frames for subsequent processing.
	
\begin{figure}[h]
		\centering
		\includegraphics[width=1\linewidth]{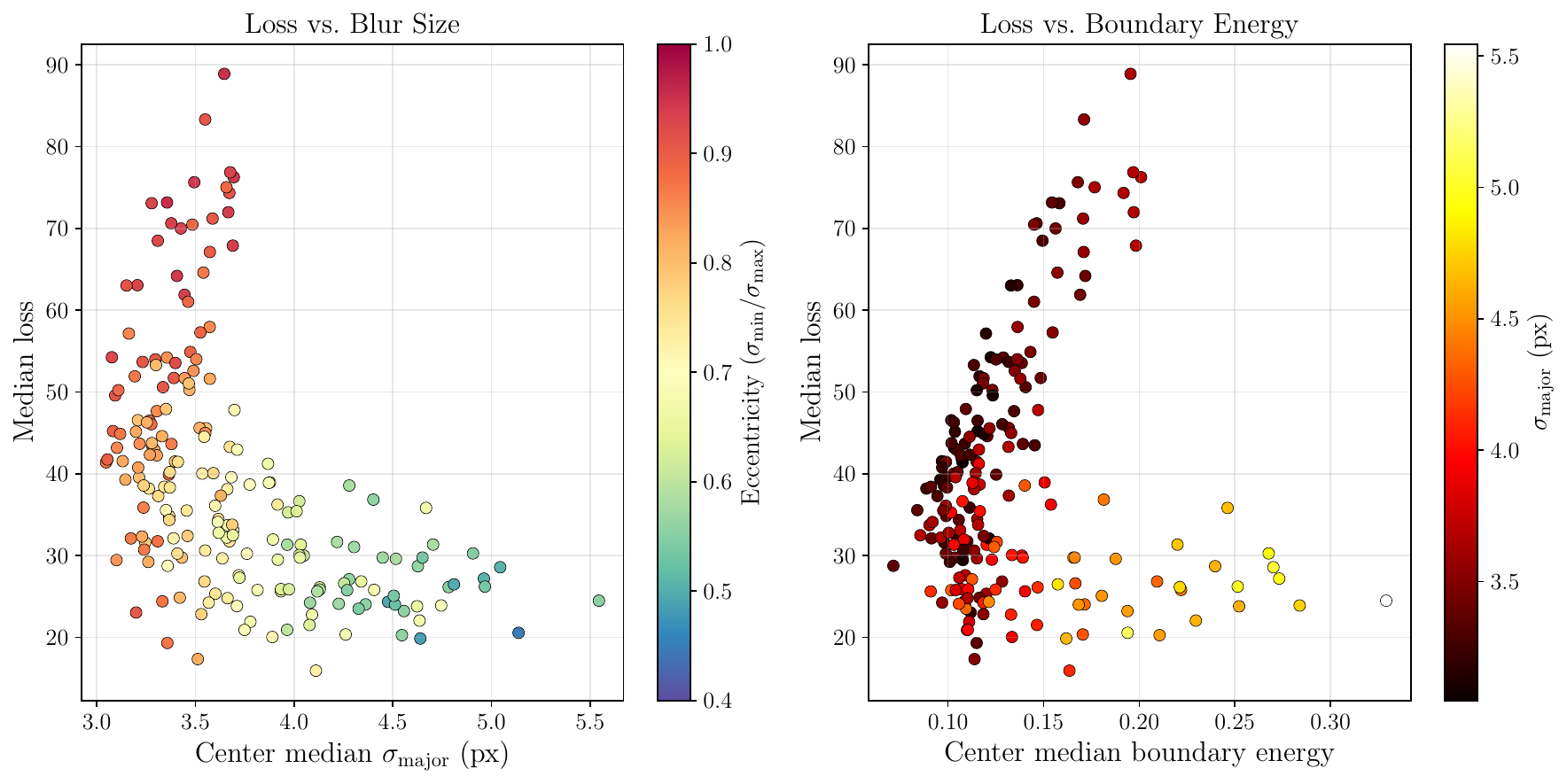}
		\caption{Cross image PSF quality analysis. Left: loss vs.\ blur size, colored by eccentricity. Right: loss vs.\ boundary energy ratio, colored by $\sigma_{\mathrm{major}}$. High boundary energy indicates PSF truncation and correlates with elevated loss, motivating the element level filtering.}
		\label{fig:exp_real2}
	\end{figure}

	\paragraph{Quality filtering.}
	Not all grid elements within a frame yield reliable estimates. We characterize each element by two metrics: the deconvolution loss from Eq.~\ref{eq:deconvolution}, and the boundary energy ratio (BE), defined as the fraction of PSF energy contained in the outer two pixel ring of the estimation window. High BE indicates that the true PSF extends beyond the estimation window and is truncated, producing biased results regardless of the loss value. Fig.~\ref{fig:exp_real2} confirms this: the right panel shows that elements with $\mathrm{BE} > 0.15$ exhibit systematically elevated loss, while the left panel reveals that low $\sigma_{\mathrm{major}}$ images (defocus dominated, high eccentricity) tend to have higher and more variable loss than motion blur dominated images. Based on these observations, we apply a two stage filter: first, elements with $\mathrm{BE} > \tau_{\mathrm{BE}}$ are removed; then, elements with loss $> \tau_L$ are removed. Connectivity is maintained via BFS from a seed element with the lowest loss in the center region.
	
\begin{figure}[h]
		\centering
		\includegraphics[width=0.6\linewidth]{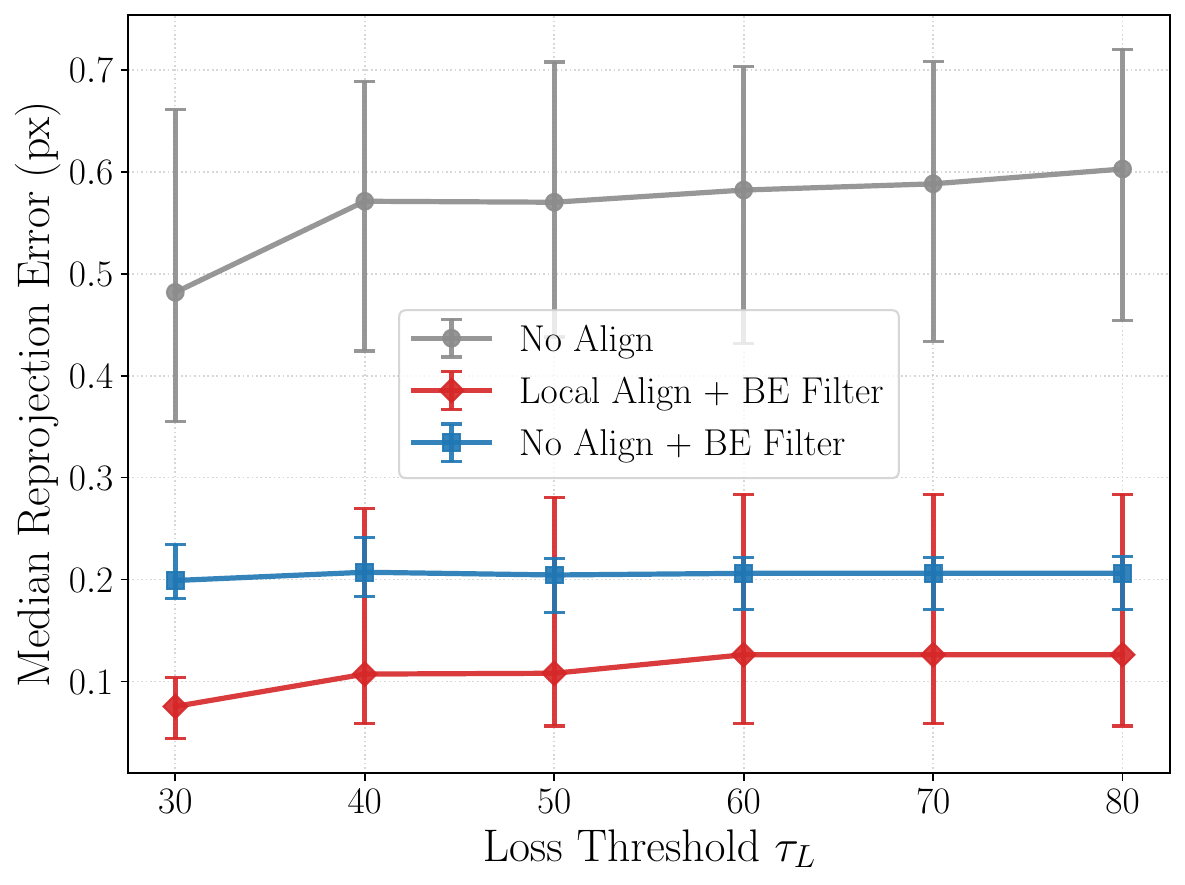}
		\caption{Median reprojection error after global alignment with bilinear bias field (Eq.~\ref{eq:bilinear_bias}). Error bars denote interquartile range. Local alignment with BE filtering (red) achieves ${\sim}\,0.08$\,px at $\tau_L{=}30$, outperforming both raw translation (gray) and BE only filtering (blue).}
		\label{fig:exp_real3}
	\end{figure}

	\paragraph{Alignment evaluation.}
	Fig.~\ref{fig:exp_real3} shows the median reprojection error after global alignment with the bilinear bias field (Eq.~\ref{eq:bilinear_bias}) at $\tau_{\mathrm{BE}} = 0.12$, evaluated across loss thresholds $\tau_L$. Three configurations are compared: raw homography translation without any filtering or local alignment (gray), BE filtered elements without local alignment (blue), and the full pipeline with both local alignment and BE filtering (red). The full pipeline achieves a median alignment error of ${\sim}\,0.08$\,px at $\tau_L = 30$, demonstrating that the combination of local alignment, quality filtering, and spatially varying bias compensation is essential for subpixel accuracy.
	
%	\paragraph{Calibration feasibility analysis.}
%	The alignment error of our method (${\sim}\,0.08$\,px, Fig.~\ref{fig:exp_real3}) is well below the reprojection error observed when evaluating parametric cameras against generic models. In Sec.~\ref{sec:global_align_verify}, using the D435I dataset of~\cite{Schops2020} (same camera model as ours), sharp images achieve a self reprojection error of $0.028$\,px (Tab.~\ref{tab:exp4}), yet the gap between parametric and generic models manifests as systematic directional bias of comparable or larger magnitude. Since the parametric model's systematic bias $\mathbf{b}(\mathbf{r})$ in Eq.~\ref{eq:error_decomposition} is deterministic and identical across all frames, it is preserved and absorbed by the generic camera model during calibration. The stochastic component of our feature estimation error averages down as $\mathcal{O}(1/\sqrt{N})$ over $N$ frames. With 42 blurry frames contributing features at diverse viewing angles, the effective per ray noise reduces to approximately $0.08 / \sqrt{42} \approx 0.012$\,px, well below the typical gap between parametric and generic models. This confirms that features extracted from blurry images by our method carry sufficient geometric precision to contribute meaningfully to generic camera calibration alongside sharp image features.
	\paragraph{Calibration feasibility analysis.}
	As shown by Sch\"{o}ps~\etal~\cite{Schops2020}, the primary advantage of 
	generic over parametric camera models lies not in lower reprojection error 
	but in the elimination of systematic directional bias (Fig.~\ref{fig:exp42}). 
	For generic calibration to benefit from additional observations, the critical 
	requirement is that the contributed features are \emph{unbiased}, \ie, their 
	errors are zero mean stochastic noise rather than systematic shifts. Our 
	alignment error of ${\sim}\,0.08$\,px (Fig.~\ref{fig:exp_real3}) is stochastic 
	by construction: the error decomposition in Eq.~\ref{eq:error_decomposition} 
	shows that the deterministic bias $\mathbf{b}(\mathbf{r})$ is absorbed by global 
	alignment, leaving only the zero mean terms $\epsilon_{\text{deconv}}$ and 
	$\epsilon_{\text{align}}$. Adding these features to the calibration increases 
	per ray observation count and angular coverage without introducing directional 
	bias, which is precisely the condition under which generic calibration improves.
	
	\section{Conclusion}
	We present the first framework for utilizing motion blurred images in generic camera calibration. Our homography parameterized local deconvolution jointly estimates feature positions and spatially varying PSFs, while the combination of local alignment, quality filtering, and bilinear bias field compensation resolves the translational ambiguity inherent in deconvolution. On real data captured with an Intel RealSense D435I, our method achieves a feature alignment error of ${\sim}\,0.08$\,px on blurry images, demonstrating that motion blurred frames can provide geometrically precise features suitable for calibration. As this work establishes a preliminary framework for a previously unexplored problem, we hope it inspires further research on topics such as motion prior integration, more robust PSF estimation, and extension to rolling shutter cameras.
	
	\clearpage
	
	\bibliographystyle{splncs04}
	\bibliography{main}
	
	\chapter*{Supplementary material}
	\appendix
	\section{Differentiable Ideal Pattern Sampling}
	
	\subsection{Pattern Definition}
	
	The calibration target consists of a periodic tiling of identical cells with period~2 in both $x$ and $y$. Each cell, centred at an even-integer lattice point, is partitioned into 16 equal angular sectors; odd-numbered sectors are white (1) and even-numbered sectors are black (0).
	
	Formally, for a point $\mathbf{q} = (q_x, q_y)$ in \emph{pattern space}, define the cell-local coordinates:
	\begin{equation}\label{eq:cell-local}
		u = q_x - 2\left\lfloor \tfrac{q_x}{2} \right\rceil, \qquad
		v = q_y - 2\left\lfloor \tfrac{q_y}{2} \right\rceil,
	\end{equation}
	where $\lfloor \cdot \rceil$ denotes rounding to the nearest integer. The local polar angle is:
	\begin{equation}\label{eq:theta}
		\theta = \operatorname{atan2}(v,\, u) \;\bmod\; 2\pi,
	\end{equation}
	and the sector index and hard pattern value are:
	\begin{equation}\label{eq:hard}
		s = \left\lfloor \frac{8\theta}{\pi} \right\rfloor \bmod 16, \qquad
		P_{\mathrm{hard}} = s \bmod 2.
	\end{equation}
	
	\subsection{Smooth Approximation via $\sin(8\theta)$}
	
	$P_{\mathrm{hard}}$ is piecewise constant, rendering direct gradient-based optimisation impossible. We observe that $\sin(8\theta)$ has exactly 16 zero-crossings in $[0, 2\pi)$, precisely at the sector boundaries $\theta = k\pi/8$. Its sign alternates between consecutive sectors, allowing the hard pattern to be expressed as $P_{\mathrm{hard}}(\theta) = H(-\sin(8\theta))$, where $H$ is the Heaviside step function.
	
	Replacing $H$ with a sigmoid $\sigma(x) = (1 + e^{-x})^{-1}$ yields a differentiable approximation:
	\begin{equation}\label{eq:soft}
		\boxed{\;
			P_\sigma(\theta) = \sigma(-\beta \sin(8\theta)),
			\;}
	\end{equation}
	where $\beta > 0$ controls the edge sharpness.
	
	\subsection{Forward Model with Analytic Brightness Solution}
	
	Let $\mathbf{p} = (p_x, p_y)$ be pixel coordinates, $\mathbf{H}$ the $3 \times 3$ homography, $h$ the point spread function (PSF), and $G$ the observed image.
	
	\paragraph{Step 1: Inverse homography mapping.}
	\begin{equation}\label{eq:inv-homo}
		\begin{pmatrix} \tilde q_x \\ \tilde q_y \\ \tilde q_w \end{pmatrix}
		= \mathbf{H}^{-1} \begin{pmatrix} p_x \\ p_y \\ 1 \end{pmatrix}, \qquad
		q_x = \frac{\tilde q_x}{\tilde q_w}, \quad
		q_y = \frac{\tilde q_y}{\tilde q_w}.
	\end{equation}
	
	\paragraph{Step 2: Cell-local coordinates and angle.}
	Cell centres $\mathbf{c} = (c_x, c_y)$ are computed by~\eqref{eq:cell-local} and \emph{detached from the computational graph}. Then:
	\begin{equation}
		u = q_x - c_x, \quad v = q_y - c_y, \quad \theta = \operatorname{atan2}(v, u).
	\end{equation}
	
	\paragraph{Step 3: Brightness model (Conditional Least Squares).}
	The predicted image $I(\mathbf{p})$ is modelled as:
	\begin{equation}
		I(\mathbf{p}) = P_\sigma(\mathbf{p})\, A(\mathbf{p}) + B(\mathbf{p}),
	\end{equation}
	where $A$ and $B$ are low-order polynomials. In each optimization closure, the coefficients for $A$ and $B$ are solved \emph{analytically} via linear least squares to match the observed intensities $G$ given the current $P_\sigma$. Consequently, $A$ and $B$ are treated as \textbf{constants} during the back-propagation of the homography $\mathbf{H}$.
	
	\paragraph{Step 4: PSF convolution and loss.}
	\begin{equation}\label{eq:loss}
		\hat I = I * h, \qquad
		\mathcal{L} = \frac{1}{N}\sum_{\mathbf{p}}\bigl(\hat I(\mathbf{p}) - G(\mathbf{p})\bigr)^2.
	\end{equation}
	
	\subsection{Gradient Derivation}
	
	As $A$ and $B$ are solved analytically and detached from the gradient flow, the derivative $\frac{\partial \mathcal{L}}{\partial \mathbf{H}}$ focuses purely on the geometric alignment of the pattern edges.
	
	\subsubsection{Loss to Predicted Image and Conv Layer}
	The gradient map $\frac{\partial\mathcal{L}}{\partial I}$ is obtained by convolving the residual $\frac{2}{N}(\hat I - G)$ with the spatially flipped PSF $\tilde{h}$.
	
	\subsubsection{Soft Pattern to Homography}
	Using the chain rule and Step 3:
	\begin{equation}
		\frac{\partial I(\mathbf{p})}{\partial P_\sigma(\mathbf{p})} = A(\mathbf{p}), \qquad
		\frac{\partial P_\sigma}{\partial \theta} = -8\beta \cos(8\theta)\;\sigma'(-\beta\sin(8\theta)).
	\end{equation}
	
	\subsubsection{Full Gradient (Assembled)}
	\begin{equation}\label{eq:full-grad}
		\boxed{\;
			\frac{\partial \mathcal{L}}{\partial H_{kl}}
			= \sum_{\mathbf{p}}
			\left[ \frac{\partial \mathcal{L}}{\partial I(\mathbf{p})} \cdot A(\mathbf{p}) \right]
			\cdot \frac{\partial P_\sigma}{\partial \theta}
			\cdot
			\left(
			\frac{\partial\theta}{\partial u}\frac{\partial q_x}{\partial H_{kl}}
			+ \frac{\partial\theta}{\partial v}\frac{\partial q_y}{\partial H_{kl}}
			\right),
			\;}
	\end{equation}
	where $\partial q_x / \partial H_{kl}$ follows from the derivative of the inverse homography $\mathbf{M} = \mathbf{H}^{-1}$ as $\frac{\partial \mathbf{M}}{\partial H_{kl}} = -\mathbf{M} \mathbf{E}_{kl} \mathbf{M}$.
	
	\subsection{Treatment of Cell Boundaries}\label{sec:cell-boundary}
	
	Cell centres $(c_x, c_y)$ are detached from the graph, justified by:
	\begin{enumerate}
		\item \textbf{Measure-zero effect:} Boundaries occupy negligible image area.
		\item \textbf{PSF smoothing:} Convolution attenuates any boundary artifacts.
		\item \textbf{Pattern continuity:} Adjacent cells carry matching values at their borders, resulting in near-zero missing gradients.
	\end{enumerate}
	
	\section{Continuous Bilinear Bias Field for Global Alignment}
	
	\subsection{Motivation: From Discrete Patches to Continuous Fields}
	
	In the initial quadrant based alignment, the grid is partitioned into four independent sections, specifically top left, top right, bottom left, and bottom right. The local drift produced by the deconvolution process was modeled as separate constant offsets for each quadrant. However, this discrete model introduces artificial jump discontinuities at the quadrant boundaries, which are physically inconsistent with the smooth nature of homography induced propagation errors. 
	
	We propose a Continuous Bilinear Bias Field that models the systematic drift across the entire grid as a low order polynomial surface. This ensures spatial continuity, provides global constraints even in feature sparse regions, and captures complex deformations such as rotation, scaling, and shear within the residual field.
	
	\subsection{Mathematical Model}
	
	For a feature point located at grid indices $(i, j)$, where $i \in [0, M-1]$ and $j \in [0, N-1]$, the bias $\mathbf{b}(i, j) = (b_x, b_y)$ is defined as a bilinear function:
	\begin{equation}\label{eq:poly-model}
		\begin{aligned}
			b_x(i, j) &= a_0 + a_1 i + a_2 j + a_3 ij, \\
			b_y(i, j) &= c_0 + c_1 i + c_2 j + c_3 ij,
		\end{aligned}
	\end{equation}
	where $\mathbf{a} = [a_0, a_1, a_2, a_3]^\top$ and $\mathbf{c} = [c_0, c_1, c_2, c_3]^\top$ are the 8 parameters governing the field. To ensure numerical stability during optimization, grid coordinates are normalized to the range $[-1, 1]$.
	
	\subsection{Iterative Alternating Optimization}
	
	The estimation of the homography $\mathbf{H}$ and the bias field parameters $(\mathbf{a}, \mathbf{c})$ is performed via an alternating iterative scheme.
	
	\paragraph{Step 1: Pose Estimation (solvePnP).}
	Given the current estimate of the bias field, the observed pixel coordinates $\mathbf{p}_{obs}$ are corrected as $\mathbf{p}_{corr} = \mathbf{p}_{obs} - \mathbf{b}(i, j)$. The global camera pose is then updated by solving the PnP problem using all corrected points across the entire grid, ensuring maximum geometric rigidity.
	
	\paragraph{Step 2: Residual Computation.}
	The reprojection error in the observation space is computed using the updated homography:
	\begin{equation}
		\mathbf{r}_k = \mathbf{p}_{obs, k} - \text{proj}(\mathbf{Q}_k; \mathbf{H}),
	\end{equation}
	where $\mathbf{Q}_k$ is the 3D world coordinate of the $k$ th feature.
	
	\paragraph{Step 3: Weighted Least Squares Fitting.}
	The parameters $\mathbf{a}$ and $\mathbf{c}$ are updated by minimizing the weighted sum of squared differences between the field and the residuals. The weights $w_k = 1 / (\text{loss}_k + \epsilon)$ are derived from the initial pattern matching stage, ensuring that high confidence features dominate the field fitting.
	
	\subsection{Normal Equations Derivation}
	
	Let $\mathbf{\phi}(i, j) = [1, i, j, ij]^\top$ be the basis vector. The objective function for the $x$ direction is defined as:
	\begin{equation}
		E(\mathbf{a}) = \sum_k w_k \left( r_{x,k} - \mathbf{\phi}(i_k, j_k)^\top \mathbf{a} \right)^2.
		\label{eq:wls_obj}
	\end{equation}
	Setting the partial derivatives with respect to $\mathbf{a}$ to zero yields the normal equations:
	\begin{equation}
		\left( \sum_k w_k \mathbf{\phi}_k \mathbf{\phi}_k^\top \right) \mathbf{a} = \sum_k w_k r_{x,k} \mathbf{\phi}_k.
	\end{equation}
	In matrix form, this is expressed as $\mathbf{M} \mathbf{a} = \mathbf{v}_x$, where $\mathbf{M}$ is a $4 \times 4$ positive definite matrix representing the weighted spatial distribution of the features. The solution $\mathbf{a} = \mathbf{M}^{-1} \mathbf{v}_x$ and the corresponding solution for $\mathbf{c}$ provide the updated field parameters.
	
\end{document}